\documentclass[lettersize,journal]{IEEEtran}
\usepackage{amsmath,amsfonts}
\usepackage{algorithm}
\usepackage{array}
\usepackage[caption=false,font=normalsize,labelfont=sf,textfont=sf]{subfig}
\usepackage{textcomp}
\usepackage{stfloats}
\usepackage{url}
\usepackage{verbatim}
\usepackage{graphicx}
\usepackage{cite}
\usepackage{booktabs}
\usepackage{multirow}
\usepackage{graphicx} 
\usepackage{hyperref}
\usepackage{makecell}
\usepackage{mathtools}
\usepackage{algpseudocode} 
\usepackage{enumitem}
\usepackage{tabularx} 
\usepackage{array}   
\newcolumntype{L}{X} 
\usepackage{orcidlink} 
\hypersetup{hidelinks}
\begin{document}

\title{\huge LandslideFM: A Foundation Model Towards Unified Landslide Perception and Understanding}

\author{\IEEEauthorblockN{Shanwen Wang, ~\IEEEmembership{Graduate Student Member,~IEEE}, Sirui Wang, Xin Sun,~\IEEEmembership{Senior Member,~IEEE}, \\Jungong Han,~\IEEEmembership{Senior Member,~IEEE}}

\thanks{This work is supported by the Science and Technology Development Fund - International Collaborative Research, Macao SAR (0001/2025/AIJ), Ministry of Science and Technology- National Key Research and Development Program of China (2025YFE0202900), Science and Technology Development Fund (FDCT-MOST), Macao SAR (0007/2025/AMJ), Science and Technology Development Fund - Basic Research of Macao SAR (0006/2024/RIA1), and Beijing Natural Science Foundation (L257005).} 
\thanks{S. Wang and X. Sun are with the Faculty of Data Science, City University of Macau, 999078, SAR Macao, China (e-mails: wangshanwen03@foxmail.com, sunxin1984@ieee.org).}
\thanks{S.R. Wang is with Data Science in Earth Observation, Technical University of Munich, Munich, 80333, Germany (e-mail: sylviaw0731@gmail.com).}
\thanks{J. Han is with the Department of Automation, Tsinghua University, Beijing, 100084, China (e-mail: jungonghan77@gmail.com).}}

\markboth{Journal of \LaTeX\ Class Files,~Vol.~14, No.~8, August~2026}
{Shell \MakeLowercase{\textit{et al.}}: A Sample Article Using IEEEtran.cls for IEEE Journals}

\maketitle

\begin{abstract}
Intelligent landslide detection is essential for disaster prevention. However, existing models primarily focus on isolated visual tasks, require extensive labeled data, and lack deep analytical and scene-understanding capabilities. To address these limitations, we propose LandslideFM, the first unified foundation model that connects visual perception tasks with contextual interpretation of landslides. Specifically, LandslideFM adapts to multiple visual tasks via a Source-Agreement Graduated Trust semi-supervised learning framework that leverages limited labeled data alongside abundant unlabeled data. Furthermore, predicted landslide regions provide explicit spatial guidance to a Vision-Language Model adapted through parameter-efficient fine-tuning, enabling comprehensive landslide reasoning. To support training and evaluation in this domain, we also present LandslideVL, the first benchmark for intelligent landslide analysis, combining visual imagery and newly introduced textual prompts. Comprehensive experiments across public datasets and LandslideVL demonstrate that LandslideFM achieves superior quantitative and qualitative performance. Most importantly, LandslideFM surpasses general-purpose VLMs in specialized landslide reasoning and scene analysis. Our code is released at \href{https://github.com/wangshanwen001/LandslideFM}{https://github.com/wangshanwen001/LandslideFM}.
\end{abstract}


\section{Introduction}
\IEEEPARstart{L}{andslides} are widespread natural disasters that threaten human lives, infrastructure, and ecosystems, causing substantial economic losses and environmental damage worldwide \cite{casagli2023landslide, ghorbanzadeh2022landslide, dille2022acceleration}. Rapid and precise landslide detection is critical for effective disaster monitoring, risk assessment, and emergency response \cite{11495246, 11550209, 11479286}. Recent advances in deep learning have driven landslide detection from convolutional neural network (CNN)-based methods toward Transformer-based architectures \cite{meng2025tlstmf, 11108280, 11227123}. This shift reflects the broader evolution from local feature extraction to long-range dependencies and global context modeling. However, existing methods typically address isolated tasks that  focus exclusively on visual perception, such as classification or segmentation. Consequently, they fail to perform higher-order analytical reasoning or incorporate broader domain knowledge into the analysis of landslide dynamics.

With the rapid development of large-scale models, new paradigms have emerged to address the aforementioned challenges. In particular, recent advances in large language models (LLMs), vision-language models (VLMs), and vision foundation models (VFMs) have attracted increasing attention from the Earth observation community \cite{10768939, hong2024spectralgpt, zhu2026foundations, zhang2026spectralx}. To support diverse downstream tasks within unified frameworks and enhance the reasoning capabilities of RS systems, a growing number of foundation models (FMs) have been developed, such as Fleximo \cite{li2026fleximo} and SpectralGPT \cite{hong2024spectralgpt}. However, most existing FMs are designed for general-purpose remote sensing (RS) applications, while their potential for specialized disaster-related tasks, such as landslide detection, remains underexplored. Consequently, despite their strong performance and generalization capabilities across broad RS tasks, these models still exhibit notable limitations in landslide scenarios, particularly in achieving precise detection and sophisticated high-level reasoning.

 \begin{figure}[!t]
\centering
\includegraphics[width=3.5in]{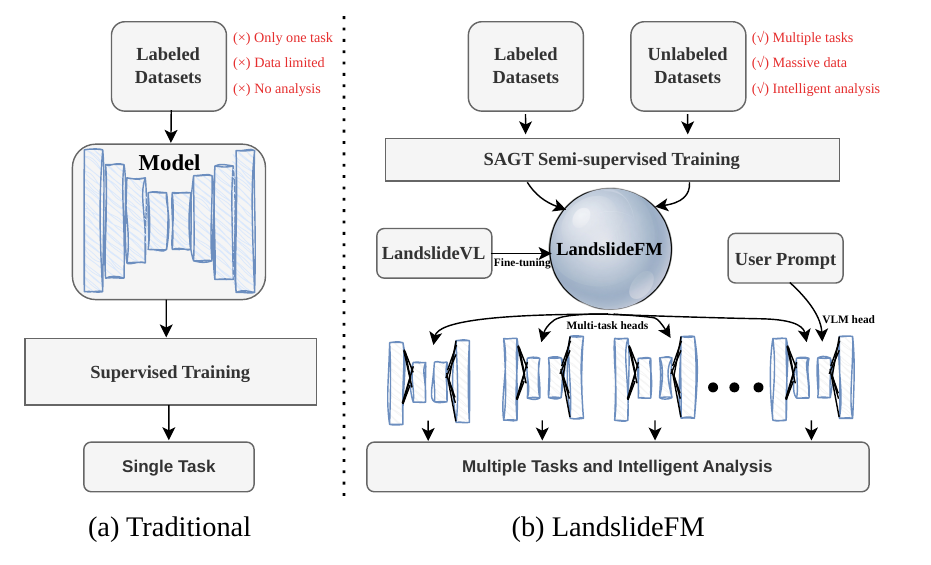}
\vspace{-0.5cm}
\caption{Comparison of traditional models (a) versus our LandslideFM (b). While traditional models are limited to specific single tasks, LandslideFM supports diverse tasks to enable intelligent, comprehensive landslide analysis.}
\label{fig: fg1}
\vspace{-0.5cm}
\end{figure}

To address these challenges and bridge this research gap, we propose LandslideFM, the first one-for-all foundation model tailored to landslide applications. As illustrated in Fig. \ref{fig: fg1}, LandslideFM unifies multiple downstream tasks within a single framework, enabling comprehensive landslide intelligent analysis. It employs a novel source-agreement graduated trust (SAGT) semi-supervised learning framework to pre-train a generalizable backbone using large-scale unlabeled landslide data. The backbone can be efficiently adapted to diverse downstream tasks, including landslide classification, detection, and segmentation. We further introduce a novel approach for integrating a VLM into LandslideFM and fine-tuning it to enhance the model's capability for landslide understanding and analysis. Beyond detecting landslides, LandslideFM can recognize their environmental context, identify nearby infrastructure, and assess potential impacts. Furthermore, to address the lack of datasets supporting image–text intelligent analysis of landslides, we introduce LandslideVL, the first manually annotated landslide dataset to combine images, structured text, and detailed textual descriptions, with all data manually annotated and verified. The main contributions of this work are summarized as follows.

\begin{enumerate}
\item We propose LandslideFM, the first unified foundation model for landslides, which performs multiple downstream tasks and intelligent analysis to fill the domain's gap for a general-purpose model.
\item We present SAGT, a semi-supervised framework that pretrains a generalizable backbone on large-scale landslide data, and further advances the landslide analysis from perception to understanding via VLM fine-tuning.
\item We introduce LandslideVL, the first benchmark for VLM-based landslide analysis, which integrates visual imagery, structured text, and detailed descriptions to support multimodal vision-language understanding.
\item Extensive experiments demonstrate the superior performance of LandslideFM over existing methods across a wide range of visual downstream tasks, as well as its capability to perform comprehensive and intelligent landslide analysis.
\end{enumerate}

The rest of this article is organized as follows: Section \ref{sec: related work} reviews the related work. In Section \ref{sec: methods}, we formally propose and analyze our LandslideFM model. Section \ref{sec: Experiment} reports the experimental results, including comparisons with state-of-the-art methods and ablation studies. Finally, Section \ref{sec: Conclusion} concludes the article and discusses future directions.

\section{related work}
\label{sec: related work}
This section reviews relevant research on classical landslide detection, summarizes recent progress in FMs in RS, and analyzes the great potential of FMs for landslide detection.

\subsection{Landslide Perception}
Landslides are widespread hazards that can cause catastrophic damage to settlements and infrastructure \cite{11197002}. Recently, deep learning has been extensively applied to rapid landslide perception, including detection, classification, and segmentation, with models evolving from CNNs to Transformer-based architectures \cite{10699390}. For example, Chen et al. \cite{10699390} proposed MSIDNet, a multiscale Swin Transformer network for InSAR-based slow-moving landslide detection, addressing key challenges in multiscale feature extraction and fusion. Tang et al.\cite{11124213} proposed SegMamba2D, a landslide detection model which adopts an encoder–decoder architecture. The model utilizes the Mamba network to extract multiscale features and employs the state space model (SSM) to reduce computational complexity while maintaining detection accuracy. Yang et al. \cite{10596327} proposed an automatic change detection method for fine coseismic landslide mapping in vegetated areas. It integrates ensemble learning with logical and spectral features from tri-temporal remote sensing images to enable automated mapping. Li et al. \cite{11489313} proposed PF-SAM, a prompt-free, parameter-efficient adaptive framework for landslide segmentation that overcomes the manual prompt dependency of traditional methods.

However, existing landslide models are generally limited to specific visual tasks like classification, detection, and segmentation. Although these models can identify landslides, they cannot perform an in-depth analysis and interpret the characteristics of landslides.


\subsection{Foundation Models in Remote Sensing}
Foundation models (FMs) are deep learning architectures that serve as foundational backbones for diverse downstream tasks \cite{awais2025foundation, wang2023internimage, wei2024stronger}. By learning general representations from large-scale datasets, FMs can be efficiently fine-tuned for specific applications \cite{chen2024internvl, wang2024sam, michalkiewicz2025not}. Numerous FMs have been developed across diverse domains, predominantly adopting the Vision Transformer as their architecture \cite{xiang2025vision, guo2025foundation, yan2025multimodal, wang2025sam, heinrich2025radiov2}.

FMs in the remote sensing (RS) domain are generally smaller than those in the general computer vision areas, placing greater emphasis on domain-specific pretraining \cite{wang2025hypersigma, hong2026foundation, bodnar2025foundation}. To overcome high annotation costs and exploit vast amounts of unlabeled data, most RS FMs rely on self-supervised learning paradigms such as masked autoencoders (MAE) \cite{li2024s2mae}. For example, SpectralGPT \cite{hong2024spectralgpt} represents a significant architectural scale-up over S2MAE and pushes this paradigm further through a 3D generative pretrained transformer (3D GPT) framework. Alliance\cite{zhao2025alliance} adopts a MAE-style masked image modeling framework for self-supervised pretraining, while further incorporating spectral-spatial-frequency joint modeling to enhance representation learning. SkySense++\cite{wu2025semantic} adopts a factorized architecture to accommodate multi-modal images acquired from diverse sensors. It employs a two-stage progressive pretraining strategy on meticulously curated datasets containing 27 million multi-modal RS images, achieving strong generalization capability. Yu et al.\cite{10768939} proposed a generative foundation model, MetaEarth, which scales image generation to a global level, exploring the creation of worldwide, multi-resolution, unbounded, and virtually limitless RS images.

Despite the rapid success of FMs in Earth observation and RS, their use in monitoring and preventing disasters such as landslides remains largely unexplored. Due to the complex environments of landslide-prone areas and the scarcity of high-quality annotations, intelligent FMs with strong generalization and multi-task capability remain in critically short supply. To address this gap, we introduce FMs to landslide analysis for the first time, establishing a novel baseline for multiple vision tasks and semantic interpretation.

\section{METHODS}
\label{sec: methods}
This section is organized as follows. Section III-A describes the main framework, Section III-B introduces the principles of the semi-supervised pre-training strategy for LandslideFM, and Section III-C presents the basic principles of the downstream adaptation strategy for LandslideFM.

\subsection{Main Framework}
\begin{figure*}[!t]
\hspace*{-0.2cm}
\centering
\includegraphics[width=7.3 in]{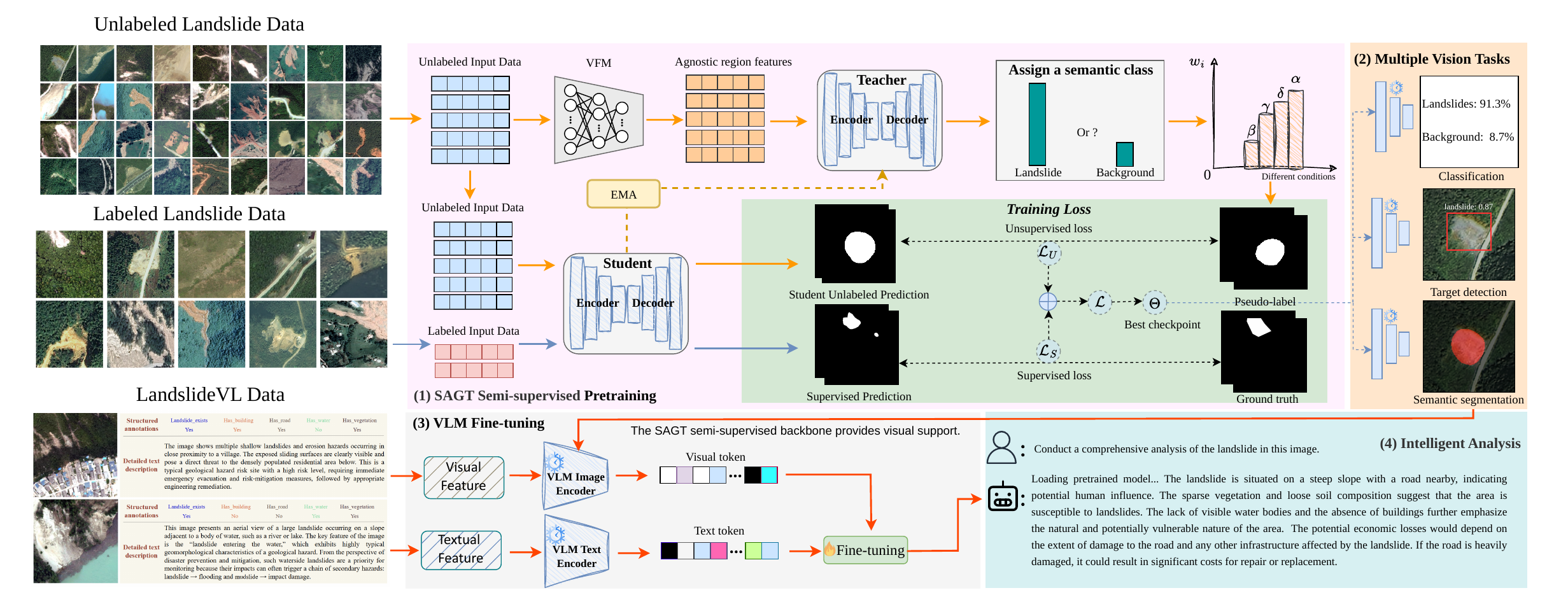}\vspace{-0.3cm}
\caption{Overall architecture of LandslideFM, which integrates SAGT semi-supervised pre-training, multiple vision tasks, VLM fine-tuning, and intelligent analysis. Yellow and blue lines indicate the flows of unlabeled and labeled data for SAGT, respectively, while orange-red lines indicate the data flow for VLM fine-tuning and the inference. Black dashed lines denote loss computation, yellow dashed lines indicate EMA updates, and blue dashed lines indicate the inference data flow for downstream vision tasks.}
\label{fig: framework}
\vspace{-0.5cm}
\end{figure*}

As illustrated in Fig. \ref{fig: framework}, the proposed LandslideFM mainly consists of two stages. First, a powerful and highly generalizable backbone is trained through our SAGT semi-supervised pretraining framework. Second, the pretrained backbone is integrated with a VLM to execute multiple vision-based downstream tasks and enable intelligent landslide analysis. During downstream tasks and intelligent analysis, all network weights  remain frozen during inference. The following subsections detail the SAGT pretraining strategy and the VLM integration with LandslideFM.

\subsection{SAGT Semi-Supervised Pretraining}

We propose the source-agreement graduated trust (SAGT) semi-supervised framework, which assigns graded trust weights based on the consistency and confidence of both the VFM and the teacher model. It transforms previously discarded inconsistent regions, where the VFM and the domain-specific teacher model disagree, into weak supervisory signals. Through a graduated response mechanism, labels with different levels of trust are assigned graded confidences rather than being indiscriminately discarded. In this subsection, we provide a detailed description of SAGT from the perspectives of data formulation, motivation, and underlying principles.

To better formulate the pre-training process of the SAGT semi-supervised method, we first define some basic concepts. We define $ D^L=\{(x_i^l,y_i)\}_{i=1}^{N_L} $ and $ D^U=\{(x_i^u)\}_{i=1}^{N_U} $ as labeled and unlabeled data. Here ${x_i^l}\in \mathbb{R}^{H\times W \times 3}$ and ${x_i^u}\in \mathbb{R}^{H\times W \times 3}$ denote labeled and unlabeled images, and ${y_i}\in \mathbb{R}^{H\times W \times K}$ is the ground truth, where $H$ and $W$ specify the height and width of the image. $N_L$ and $N_U$ denote the numbers of labeled and unlabeled images, respectively, where $N_L << N_U$. As shown in  Fig. \ref{fig: framework}, the supervised loss $\mathcal L_S$ constrains the supervised predictions of the student on labeled data and ground truth. Unsupervised loss $\mathcal L_U$ (see Eq. \ref{eq: SAGT}) constrains the consistency between the student's predictions on unlabeled data and pseudo-labels. The overall loss function $\mathcal L$ summarizes supervised $\mathcal L_S$  and unsupervised $\mathcal L_U$ losses as follows:
\begin{equation}
\label{eq: loss}
\mathcal L=\mathcal L_S+\mathcal L_U =\frac{1}{N_L}{\sum_{ {i=1}}^{N_L}\mathcal L_{CE}(p_i^l,y_i)}+ \mathcal L_U,
\end{equation}
\noindent where $p_i^l$ is the prediction of labeled image $x_i^l$, and $\mathcal L_{CE}$ is cross-entropy loss. The teacher model is updated using an exponential moving average (EMA)\cite{11612938}.

The key to improving pseudo-label quality in LandslideFM lies in generating pseudo-labels through the SAGT. The detailed procedure is as follows. First, the weakly augmented unlabeled image $x^{u,weak}_i$ is fed into the VFM to generate a set of class-agnostic region masks:
\begin{equation}
M_i = \{m^1_i, m^2_i, ... , m^j_i\} = \text{VFM($x^{u,weak}_i$)},  
\end{equation}
where $m^j_i$ denotes the $j^{th}$ class-agnostic mask region generated by the VFM for unlabeled image $i$. The VFM can generate an arbitrary number of categories of class-agnostic masks, whereas landslide datasets typically contain only two classes: landslide and background. We then use a landslide teacher model to identify valid masks and constrain them to these two classes.
The predicted class $y^{u,t}_i$ and confidence score $c^k_i$ of the teacher for each pixel are given by:
\begin{equation}
p^{u,t}_i(x^{u,weak}_i) = Softmax(Teacher(x^{u,weak}_i)).  
\end{equation}
\begin{equation}
y^{u,t}_i = \mathop{\arg\max}_{k \in [0,1]}(p^{u,t}_i(x^{u,weak}_i)),
\end{equation}
\begin{equation}
c^k_i=\mathop{\max}_{k \in [0,1]}(p^{u,t}_i(x^{u,weak}_i)),
\end{equation}
where $k$ indexes the classes, with 0 denoting landslide and 1 denoting background. However, prediction inconsistencies often arise between the VFM and the teacher model. Previous methods, such as UFFM \cite{wang2026bridging}, address this by measuring prediction consistency and discarding regions below a predefined threshold. Crucially, these inconsistent regions often correspond to hard-to-segment areas, meaning that discarding their pseudo-labels hinders the student model from learning challenging areas. To address this issue, SAGT assigns graduated trust weights based on the prediction consistency and confidence of VFM and the teacher model, thereby transforming previously discarded inconsistent regions into weak supervisory signals. With SAGT, the pseudo-label loss function simplifies to:
\begin{equation}
\label{eq: SAGT}
\mathcal L_U = \frac{{\sum_{ {i=1}}^{N_U}w_i\mathcal L_{CE}(p_i^{u,s},y_i^{u,t})}}{\sum_{ {i=1}}^{N_U}w_i}.
\end{equation}
Here, $w_i$ is the core weight of the graduated trust weighting mechanism, $p^{u,s}_i$ is student predictions for unlabeled data $x_i^u$, and $y^{u,t}_i$ is the predicted class of teacher model. The Eq. \ref{eq: SAGT} represents the pseudo-label unsupervised loss function after SAGT weighting. The core principle of the graduated trust weights  $w_i$ is as follows:
\begin{equation}
\label{eq: w_i}
w_i=
\begin{cases}
\alpha, & if\quad S_i=1\ \&\ T_i=1 \ \&\ A_i=1; \\
\beta, & if\quad S_i=1 \ \&\ T_i=1 \ \&\ A_i=0; \\
\gamma, & if\quad S_i=1 \ \&\ T_i=0; \\
\delta, & if\quad S_i=0 \ \&\ T_i=1; \\
0, & otherwise.
\end{cases}
\end{equation}
In the above equation, $\alpha$, $\beta$, $\gamma$, and $\delta$ are all constants, and their specific values can be found in the experimental section \ref{sec: Experiment Settings}.  $S_i$ indicates the reliability of the prediction made by the Segment Anything Model (SAM), $T_i$ denotes whether the teacher model yields a high-confidence prediction, and $A_i$ represents the class consistency between SAM and the teacher model. The definition of $T_i$ is given as $c^k_i > \tau$, and $\tau$ is the pseudo-label threshold. $S_i$ is defined as follows:
\begin{equation}
p^{u,v}_i(x^{u,weak}_i) = Softmax(VFM(x^{u,weak}_i)),  
\end{equation}
\begin{equation}
c^k_{i,v}=\mathop{\max}_{k \in [0,1]}(p^{u,v}_i(x^{u,weak}_i)),
\end{equation}
\begin{equation}
S_i=1*(c^k_{i,v}\ge\tau).
\end{equation}
The core idea of the above equation is to assign graded trust weights to regions where the VFM and the teacher model have inconsistent predictions, rather than directly discarding them. 

\subsection{Downstream Vision Tasks}

To apply the pre-trained backbone to downstream vision tasks, an input image $I \in \mathbb{R}^{H \times W \times 3}$ is first processed by the semi-supervised feature extractor $f_{\Theta}$ , parameterized by $\Theta$, to extract high-level semantic features $F$. These features are subsequently fed into task-specific heads $H$ to generate predictions for semantic segmentation, image classification, and object detection. Finally, final task outputs are obtained via corresponding post-processing operations, such as argmax, softmax, or threshold-based filtering with non-maximum suppression (NMS)\cite{ren2016faster}. The overall process is formulated as follows:
\begin{equation}
F={f_{\Theta}}(I, \Theta),
\end{equation}
\begin{equation}
Y_t=D_t(H_t(F)), t\in \{cls,det, seg\}.
\end{equation}
Here $H_t$ denotes the task-specific head for task $t$, and $D_t$ represents the corresponding post-processing operations for each task, such as decoding and NMS.

Specifically, for the classification head, a softmax function generates a probability distribution over all classes, and argmax selects the class with the highest probability.
\begin{equation}
P_{cls} = softmax(H_{cls}(F)),
\end{equation}
\begin{equation}
Y_{cls} = \mathop{ \arg \max}_{k \in [1,K]}(P_{cls}).
\end{equation}

For the object detection task, since multiple overlapping bounding boxes may be generated for the same object, NMS is applied as the final post-processing step. The NMS suppression procedure first ranks the detected bounding boxes according to their confidence scores and retains the highest-scoring box $B_i$ for the same landslide instance. If another bounding box $B_j$ of the same class satisfies the following condition, the two boxes are considered duplicates, and the lower-scoring box $B_j$ is suppressed.
\begin{equation}
IoU(B_i,B_j)=\frac{|B_i \cap B_j|}{|B_i \cup B_j|} \geq \tau_{NMS},
\end{equation}
where $\tau_{NMS}$ denotes the overlap threshold for NMS and is set to 0.45 by default in this study.

For the semantic segmentation head, the model performs pixel-wise classification as follows:
\begin{equation}
p_{seg}(x,y) = softmax(H_{seg}(F(x,y))),
\end{equation}
\begin{equation}
y_{seg}(x,y) = \mathop{ \arg \max}_{k \in [1,K]}(p_{seg}(x,y)),
\end{equation}
where $(x,y)$ denotes the coordinates of a pixel, and $Y_{seg}$ is the final segmentation result containing all pixels. 

Overall, LandslideFM employs a unified backbone pre-trained in a semi-supervised manner to support various downstream vision tasks.

\subsection{VLM Fine-Tuning and Intelligent Landslide Analysis}
This section presents our framework for intelligent landslide analysis. Rather than a standalone image classification model, the proposed method is a cascaded architecture that integrates a pretrained landslide vision model, a vision-language model (VLM), and LoRA-based fine-tuning. Specifically, the framework is structured as follows. The backbone, pretrained via the SAGT semi-supervised learning framework, firstly identifies and localizes landslide-affected areas in RS imagery. Next, the original RS image, together with a visualization overlaying the detected regions, are jointly fed into the LoRA-fine-tuned VLM. With exact spatial information of the landslides, the model can conduct a comprehensive analysis of the surrounding terrain, roads, buildings, water bodies, vegetation, and other potential influencing factors. The entire process is detailed as follows.

The original image $I$ and segmentation result $Y_{\text{cls}}$ are individually divided into patches and mapped to a unified sequence of visual tokens $[v_1, \dots, v_n]$. The text prompt is tokenized into $[t_1, \dots, t_m]$.  These visual and textual tokens are then concatenated into a combined sequence $x = [v_1, \dots, v_n, t_1, \dots, t_m]$ and fed into the VLM's self-attention module:
\begin{equation}
\text{Attention}(Q,K,V)=\text{softmax}(\frac{QK^{\mathrm{T}}}{\sqrt{d_k}})V,
\end{equation}
where $Q=xW^Q$, $K=xW^K$, and $V=xW^V$. The $W^K$ and $W^V$ matrices, as well as the parameter matrices in subsequent equations, are  learned during VLM pretraining. The VLM uses multi-head attention, allowing individual attention heads to focus on distinct image-text correspondences,
\begin{equation}
\text{MultiHead}(x)=\text{Concat}(h_1,...,h_h)W^Y,
\end{equation}
where $h_i$ denotes the $i$-th attention head. The language head then outputs the conditional probability distribution of the next token at each position:
\begin{equation}
p_\theta(t_i\mid t_{<i}, v_{n})=\text{softmax}(W_{\text{vocab}}h_i),
\end{equation}
where $t_i$ denotes the $i$-th text token to be predicted,  $t_{<i}$ denotes the sequence of tokens preceding position $i$, and $\theta$ represents the VLM's frozen base parameters. As illustrated in Fig. \ref{fig: framework}, LoRA fine-tuning is applied to the intelligent reasoning component of our model. For each target linear layer $l$ in the VLM, LoRA introduces a low-rank update:
\begin{equation}
\Delta \theta_l= W_lx+\Delta W_lx=W_lx+\frac{\alpha}{r}BAx, 
\end{equation}
where $W_l$ denotes the parameters of the frozen VLM layer, $\Delta W_l$ represents the low-rank weight update introduced by LoRA, and $\Delta \theta_l$ denotes the output after the LoRA update. Specifically, $\frac{\alpha}{r}$ is the scaling factor, and only A and B are trainable. The loss function for LoRA fine-tuning is defined as:
\begin{equation}
\mathcal L_{Lora} = - \frac{1}{|\tau|}\sum_{i\in \tau}logp_ {\Delta \theta}(t_i\mid t_{<i}, I, O),
\end{equation}
where $\tau$ denotes the manually annotated text, $i$ is the index of the corresponding token, $I$ is the original image, and $O$ represents the original image $I$ overlaid with the predicted segmentation result $Y_{\text{seg}}$.

Following LoRA training, the learned low-rank parameters are integrated directly into the base VLM  model.
\begin{equation}
p(a_1,..., a_K|,I,O,Q)=\prod_{k=1}^K p\Delta \theta(a_k\mid a_{<k},I,O,Q),
\end{equation}
where $a_k$ denotes the $k$-th answer token, $\prod$ represents the product operator, and $Q$ refers to the prompt which may be user-specified or predefined.

In this section, we outlined the structural design of LandslideFM, covering SAGT semi-supervised pretraining, downstream computer vision tasks, and intelligent landslide analysis. Next, we introduce the LandslideVL benchmark in Section~\ref{sec: LandslideVL} and validate our model's effectiveness in Section~\ref{sec: Experiment}.

\section{Landslide Analysis Benchmark: LandslideVL}
\label{sec: LandslideVL}
Existing landslide datasets focus primarily on computer vision tasks like semantic segmentation and lack textual annotations. Consequently, they are ill-suited for directly training intelligent landslide analysis models. We address this gap by building LandslideVL, a vision-language dataset for landslide analysis, which is curated and textually annotated representative remote sensing images from multiple existing landslide datasets. The detailed construction pipeline is presented below.

The landslide datasets selected in this study include Bijie Landslide (BJL) dataset \cite{ji2020landslide}, Global Very High Resolution Landslide Mapping (GVLM) dataset \cite{zhang2023cross}, and Jiuzhai valley landslide dataset \cite{xu2024cas}. The BJL dataset provides annotated optical satellite imagery from Bijie, Guizhou, a high-risk landslide region in China. It consists of 770 landslide and 2,003 non-landslide images. The GVLM dataset is the first large-scale, open-source, ultra-high-resolution (0.59 meters) landslide mapping dataset. It consists of 17 landslide sub-datasets from various geographic regions, each containing a pair of bi-temporal images and the corresponding landslide masks. The dataset covers a total area of 163.77 square kilometers and spans regions across Asia, Africa, North America, South America, Europe, and Oceania. To accommodate GPU memory constraints, we cropped these images to $512 \times 512$ pixels, obtaining 1,986 patches.
The Jiuzhai Valley landslide dataset is a high-quality landslide subset of the CAS Landslide Dataset\cite{xu2024cas}. It consists of high-resolution RGB imagery acquired by unmanned aerial vehicles (UAVs) in the Jiuzhai Valley region and contains 1,752 high-resolution landslide images. We then merged high-quality samples from these aforementioned processed datasets, including landslide images and easily confusable non-landslide images. From these samples, we selected 4,660 representative instances and supplemented them with textual annotations. The detailed information of the selected datasets can be found in Table \ref{tab: landslidevl_sources}, including geographic locations, acquisition dates, and spatial resolutions. 

 \begin{figure}[!t]
\centering
\includegraphics[width=3.5in]{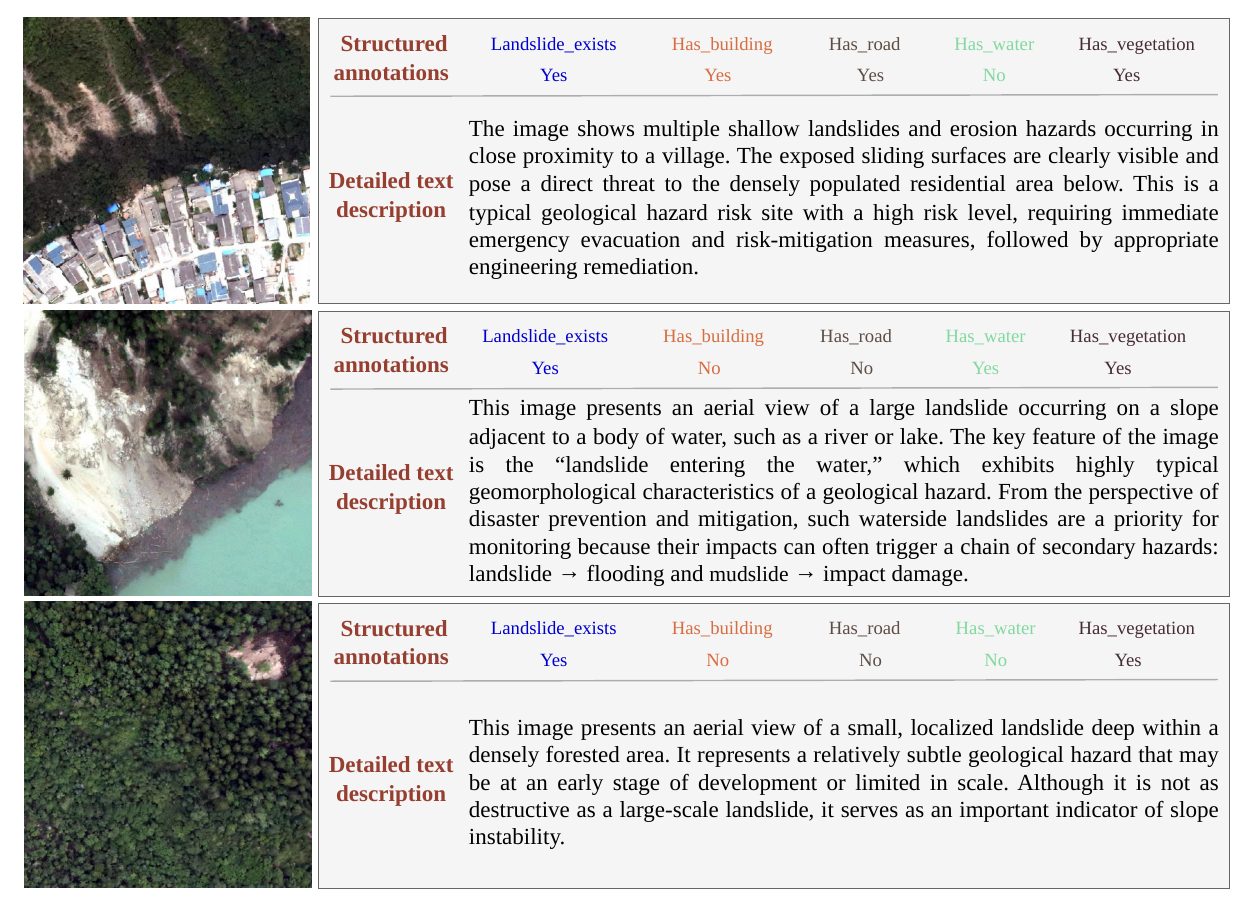}
\vspace{-0.5cm}
\caption{Examples of textual annotations in LandslideVL. Each landslide image is paired with a structured text label and a detailed textual description.}
\label{fig: LandslideVL}
\vspace{-0.5cm}
\end{figure}

\begin{table*}[!t]
\centering
\caption{Data sources and composition of LandslideVL. All textual annotations were manually provided by human annotators and underwent multiple rounds of validation.}
\label{tab: landslidevl_sources}
\begingroup
\small
\setlength{\tabcolsep}{5pt}
\renewcommand{\arraystretch}{1.18}
\begin{tabularx}{\textwidth}{@{}>{\raggedright\arraybackslash}p{0.19\textwidth}
    *{3}{>{\raggedright\arraybackslash}X}@{}}
\toprule
\textbf{Property} &
\textbf{BJL}~\cite{ji2020landslide} &
\textbf{GVLM}~\cite{zhang2023cross} &
\textbf{Jiuzhai Valley}~\cite{xu2024cas} \\
\midrule
Geographical coverage & Bijie, Guizhou, China & 17 regions across six continents & Jiuzhai Valley, Sichuan, China \\
\addlinespace[2pt]
Image acquisition period & May--August 2018 & 2010--2021 &
August--September 2017 \\
\addlinespace[2pt]
Imagery source & TripleSat satellite & Google Earth imagery & UAV imagery \\
\addlinespace[2pt]
Source data types & Optical RGB imagery & Bitemporal optical imagery & Optical RGB imagery \\
\addlinespace[2pt]
Spatial resolution (m) & 0.8 & 0.59 & 0.5 \\
\addlinespace[2pt]
Original data volume & 2,773 image patches & 17 bitemporal image pairs & 1,752 image patches \\
\midrule
Image preparation & None & Cropping to $512\times512$ pixels &
None \\
\addlinespace[2pt]
Samples before selection & 2,773 patches & 1,986 patches & 1,752 patches \\
\addlinespace[2pt]
Selected for LandslideVL & \textbf{2,267} & \textbf{1,298} & \textbf{1,095} \\
\addlinespace[2pt] Selection ratio &
81.75\% & 65.36\% & 62.50\% \\
\midrule
\multicolumn{4}{@{}l@{}}{\textbf{Total: 4,660 selected samples. All samples were manually annotated and calibrated through multiple rounds by human annotators.}} \\
\bottomrule
\end{tabularx}
\par\vspace{4pt}
\endgroup
\end{table*}
 
As shown in Fig. \ref{fig: LandslideVL}, the structured textual annotations in LandslideVL include landslide\_exists, has\_building, has\_road, has\_water, has\_vegetation. The selection of these five binary scene attributes aims to provide a concise, yet comprehensive description of landslide scenes, covering three complementary aspects: landslide occurrence, anthropogenic factors, and surrounding environmental context. Specifically, landslide\_exists serves as the fundamental indicator to determine the presence of a landslide and provides the primary semantic information for the recognition of landslides. has\_building describes important man-made structures located within or near the affected area. Since many landslides occur in remote or undeveloped regions where large human-made structures are absent, roads may represent the primary infrastructure exposed to landslide impacts. Landslides can block or damage roads, making has\_road an important attribute to characterize potential infrastructure disruption. Water bodies are also important contextual features and has\_water is used to identify the presence of rivers or other water bodies within or near the landslide area. Vegetation, in turn, provides useful information on land cover and slope-surface conditions. Changes or disturbances in vegetation may also serve as observable indicators associated with slope instability or landslide activity. In addition, each image is accompanied by a detailed textual description. 

Notably, all data in LandslideVL have been manually annotated and calibrated over multiple rounds by paid human annotators, rather than automatically annotated by AI. Our LandslideVL dataset will be publicly available on Hugging Face and can be freely downloaded for further research \href{https://huggingface.co/datasets/fluorites/LandslideVL}{https://huggingface.co/datasets/fluorites/LandslideVL}.

\section{Experiments}
\label{sec: Experiment}

To thoroughly validate the effectiveness of the proposed LandslideFM, we conduct extensive experiments on several public landslide datasets for different downstream vision tasks. Additionally, we conducted intelligent landslide analysis experiments on our newly proposed LandslideVL benchmark. We first introduce the experimental evaluation datasets and evaluation metrics, followed by comprehensive quantitative and qualitative comparisons with SOTA methods. Finally, we perform ablation studies to analyze the contribution of each component in the proposed framework. 

\subsection{Datasets and Experimental Settings}
\label{sec: Experiment Settings}
We benchmark landslide-related visual perception tasks across the aforementioned BJL \cite{ji2020landslide}, GVLM \cite{zhang2023cross}, and LandslideVL datasets. Additionally, our LandslideVL dataset is leveraged to evaluate high-level intelligent landslide analysis.

\begin{figure}[!t]
\centering
\includegraphics[width=3.3in]{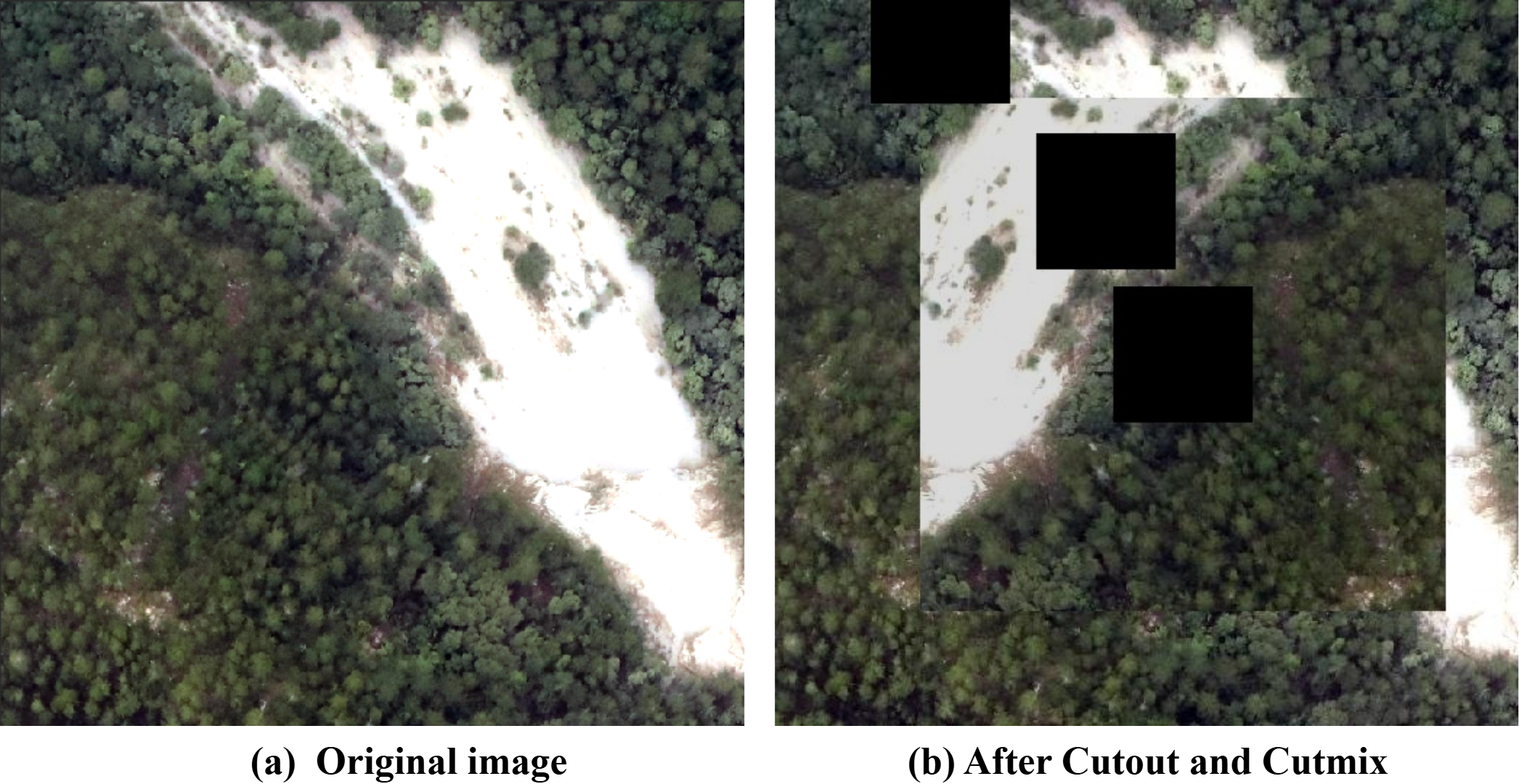}
\vspace{-0.0cm}
\caption{Cutout and CutMix disrupt the semantic consistency of landslide regions.}
\label{fig: cutoutandmix}
\vspace{-0.5 cm}
\end{figure}

Our model is implemented in PyTorch and trained on one NVIDIA RTX H100. We adopt DINOv2\_base \cite{oquab2023dinov2} as the teacher-student backbone network and utilize SAM 3 \cite{carion2026sam}  as the VFM. The $\alpha$, $\beta$, $\gamma$, and $\delta$ values for the semi-supervised pre-training of SAGT are set to 1.0, 0.3, 0.7, and 0.5, respectively. We discuss the selection of these hyperparameters in detail in the ablation study.
For the visual tasks, LandslideFM is trained separately on the BJL, GVLM, and LandslideVL datasets for 50 epochs each. For intelligent analysis, it is fine-tuned on LandslideVL for only 5 epochs. During pre-training, weak augmentation primarily involves geometric transformations, including random flipping and scaling, whereas strong augmentation includes photometric transformations, Gaussian blurring, and other perturbations. Although strong augmentation techniques such as Cutout and CutMix have shown promising results across various applications, they are not well suited to landslide segmentation. As shown in Fig. \ref{fig: cutoutandmix}, these augmentation strategies can disrupt the semantic consistency of landslide regions, which is essential for accurate identification. As landslides are spatially continuous and dynamic phenomena, their identification relies heavily on the surrounding environmental context. We therefore avoid region-cutting augmentations such as CutMix.

\subsection{Evaluation Metrics}
We evaluate our model across multiple downstream vision tasks, starting with classification using Precision, Recall, and F1-score (F1).
\begin{equation}
\label{eq: R}
Recall=\frac{TP}{TP+FN},
\end{equation}
\begin{equation}
\label{eq: P}
Precision=\frac{TP}{TP+FP},
\end{equation}
\begin{equation}
\label{eq: F1}
F1=\frac{2 \times Recall \times Precision}{Recall+Precision},
\end{equation}
where $TP$ and $TN$ denote the numbers of correctly predicted positive and negative pixels, respectively, while $FP$ and $FN$ denote the numbers of incorrectly predicted positive and negative pixels, respectively. For object detection, we additionally adopt Intersection over Union (IoU)\cite{ren2016faster} as an evaluation metric,
\begin{equation}
\label{deqn_ex1a}
IoU=\frac{TP}{TP+FN+FP}.
\end{equation}
For semantic segmentation, we additionally employ Cohen’s Kappa as an evaluation metric, defined as follows:
\begin{equation}
\label{deqn_ex1a}
OA=\frac{TP+TN}{TP+TN+FP+FN},
\end{equation}
\begin{equation}
\small
\label{deqn_ex1a}
PRE=\frac{(TP+FN)(TP+FP)+(TN+FN)(TN+FP)}{(TP+TN+FP+FN)^2},
\end{equation}
\begin{equation}
\label{deqn_ex1a}
Kappa=\frac{OA-PRE}{1-PRE}.
\end{equation}
Since all landslide datasets contain two classes, i.e., landslide and background, we average the aforementioned metrics over both classes. Accordingly, we report mean Precision (mPrecision), mean Recall (mRecall), mean IoU (mIoU), and mean F1-score (mF1). Taking mIoU as an example, it is computed as follows:
\begin{equation}
\label{deqn_ex1a}
mIoU=\frac{\sum_{i=1}^K{IoU_i}}{K},
\end{equation}
where $IoU_i$ means the IoU of the $i^{th}$ class and $K$ is the number of classes. Furthermore, to avoid ambiguity and ensure the reliability of the data, we specify that mF1 is still computed from mPrecision and mRecall according to Eq. \ref{eq: F1}, rather than by simply averaging the F1 scores across individual classes.

To evaluate the quality of free-text descriptions, we employ three metrics: BLEU\cite{papineni2002bleu}, ROUGE-L\cite{lin2004rouge}, and Word Overlap\cite{rajpurkar2016squad}. BLEU measures n-gram-level\cite{4766902} agreement between generated and reference texts. It is based on the clipped n-gram precision $p_n$ in Eq.~\ref{eq: p_n}, with a brevity penalty (BP) in Eq.~\ref{eq: bp} to penalize overly short outputs. The overall BLEU score is then computed as the geometric mean over $n$-gram orders up to four as defined in Eq.~\ref{eq: BLEU}.
\begin{equation}
\label{eq: p_n}
p_n=\frac{\sum_{g\in G_n(H)}min(cnt_H(g),cnt_R(g))}{\sum_{g\in G_n(H)}cnt_H(g)},
\end{equation}
where $R$ and $H$ denote the token sequences of the reference and generated texts, respectively. $G(\cdot)$ denotes the multiset of all n-grams extracted from a sequence, while $cnt_H(g)$ and $cnt_R(g)$ are the occurrence counts of n-gram $g$ in the generated and reference texts.
\begin{equation}
\label{eq: bp}
BP = \begin{cases}
1,   & if \ \ \ l_H>l_R,  \\
exp(1-\frac{l_R}{l_H}), & if \ \ \  l_H<l_R
\end{cases},
\end{equation}
where $l_R=|R|$ and $l_H=|H|$.
\begin{equation}
\label{eq: BLEU}
\text{BLEU}=BP\cdot exp(\frac{1}{4}\sum_{n=1}^4 \ln p_n).
\end{equation}

ROUGE-L evaluates the structural coherence between generated and reference texts based on the Longest Common Subsequence (LCS). Using the LCS length, ROUGE-L precision $\text{ROUGE-L}_P$, recall $\text{ROUGE-L}_R$, and F1 score $\text{ROUGE-L}_{F1}$ are calculated as follows:
\begin{equation}
\text{ROUGE-L}_P=\frac{LCS(R,H)}{|H|},
\end{equation}
\begin{equation}
\text{ROUGE-L}_R=\frac{LCS(R,H)}{|R|},
\end{equation}
\begin{equation}
\text{ROUGE-L}_{F1}=\frac{2*LCS(R,H)}{|R|+|H|}.
\end{equation}

Word Overlap measures the bidirectional coverage of unique words between the generated text and the reference text at the set level, and is reported in terms of precision $P_{WO}$, recall $R_{WO}$, and F1 score $F1_{WO}$. The calculation follows Eqs. (\ref{eq: R}) - (\ref{eq: F1}), except that text replaces images as the evaluation object.

\subsection{Comparison with SOTA Methods}
\subsubsection{Classification}
Our approach uses semi-supervised pre-training, whereas the baseline classifiers do not. To facilitate a fair comparison, all methods use the same labeled-data split, with 80\% allocated to training and the remaining 20\% to testing.

\begin{table}[h]
\caption{Comparison of classification results with other methods on landslide datasets. the best results are highlighted in bold.}
\label{tab: classification}
\centering
\begin{tabular}{c|cccc}
    \toprule
    \textbf{Datasets} & \textbf{Method} & \textbf{Precision} & \textbf{Recall} & \textbf{F1} \\
    \midrule
    \multirow{4}{*}{BJL}
        & ResNet101\cite{he2016deep} & 96.08 & 93.33 & 94.69 \\
        & DenseNet\cite{huang2017densely} & 95.12 & 92.18 & 93.63 \\
        & ViT\cite{han2022survey} & 96.41 & 95.45 & 95.93 \\
        & DINO\_v2\cite{oquab2023dinov2} & 96.25 & 95.81 & 96.03  \\
        & LandslideFM (1/8) & 96.74 & 95.96 & 96.35 \\
        & \textbf{LandslideFM} (Full) & \textbf{99.08} & \textbf{98.70} & \textbf{98.89} \\
    \midrule
    \multirow{4}{*}{GVLM}
       & ResNet101\cite{he2016deep} & 84.02 & 76.89 & 80.30 \\
        & DenseNet\cite{huang2017densely} & 85.72 & 80.11 & 82.82 \\
        & ViT\cite{han2022survey} & 85.67 & 88.25 & 86.94 \\
         & DINO\_v2\cite{oquab2023dinov2} & 85.48 & 87.90 & 86.67  \\
         & LandslideFM (1/8) & 86.22 & 88.47 & 87.33 \\
        & \textbf{LandslideFM} (Full) & \textbf{90.81} & \textbf{91.30} & \textbf{91.06} \\
    \bottomrule
\end{tabular}
\end{table}

Table \ref{tab: classification} presents the classification results. On the BJL dataset, all evaluated methods demonstrated strong performance, with precision, recall, and F1-score all exceeding 90\%. LandslideFM achieved further gains, notably reaching 99.08\% precision. On the GVLM dataset, ViT achieved precision comparable to traditional CNN-based methods while significantly improving recall. LandslideFM outperformed all competing models across all metrics on the GVLM dataset. For a fair comparison, LandslideFM was trained using the same volume of labeled data as the baseline models. In fact, as shown in Table \ref{tab: classification}, our experiments show that our method can already outperform the competing methods on the evaluation metrics when using only approximately one-eighth of the data as labeled data and treating the remaining data as unlabeled. This is because LandslideFM leverages SAGT semi-supervised pre-training  to make more effective use of limited labeled data, allowing it to significantly outperform traditional fully supervised methods in label-scarce scenarios.

\subsubsection{Object Detection}

Similar to the classification setup, all object detection models are trained on 80\% of the labeled data and tested on 20\% to maintain a fair baseline against algorithms that lack semi-supervised support. Because the BJL and GVLM datasets lack explicit bounding-box annotations, we derive ground-truth bounding boxes directly from their ground-truth masks. The thresholds for the object detection boxes were set to 0.5 and 0.75 in the experiments. For evaluation, the maximum geometric IoU between each GT box and all predicted boxes was calculated, and the resulting values were then averaged over all GT boxes.

\begin{table}[h]
\caption{Comparison of object Detection results with other methods on landslide datasets. the best results are highlighted in bold.}
\label{tab: Detection}
\centering
\begin{tabular}{c|ccccc }
    \toprule
    \textbf{Datasets} & \textbf{Method} & \textbf{Precision} & \textbf{Recall} & \textbf{F1}  & \textbf{IoU}\\
    \midrule
    \multirow{4}{*}{BJL}
       & Faster R-CNN\cite{ren2016faster} & 55.88 & 74.03 & 63.69 & 65.91 \\
        & DETR\cite{carion2020end} & 57.87 & 74.04 & 64.96 & 65.97 \\
        & DINO\_v2\cite{oquab2023dinov2} & 73.33 & 79.08 & 76.10 & 68.69\\
         & LandslideFM (1/8) & 72.16 & 77.62 & 74.79 & 67.65 \\ 
        & \textbf{LandslideFM} (Full) & \textbf{75.16} & \textbf{80.13} & \textbf{77.56} & \textbf{69.85} \\
    \midrule
    \multirow{4}{*}{GVLM}
        & Faster R-CNN\cite{ren2016faster} & 37.88 & 50.32 & 43.22 &  48.66 \\
         & DETR\cite{carion2020end} & 43.45 & 52.61 & 47.60 & 50.84 \\
        & DINO\_v2\cite{oquab2023dinov2} & 45.79 & 56.57 & 50.61 & 53.67 \\
          & LandslideFM (1/8) & 48.39 & 57.92 & 52.73  & 54.22 \\
          & \textbf{LandslideFM} (Full) & \textbf{57.68} & \textbf{61.64} & \textbf{59.59} & \textbf{57.72}\\
    \bottomrule
\end{tabular}
\end{table}

Table \ref{tab: Detection} presents the results for the object detection task. Compared to the classification results, all methods exhibit a decline across all evaluation metrics. Furthermore, because the bounding-box annotations are simulated from ground-truth masks, the overall performance of most algorithms remains limited. Specifically, none of the evaluated models achieves an IoU exceeding 70\% on BJL or 60\% on GVLM. However, this setup closely reflects real-world scenarios where precise annotations are unavailable. Despite these challenges, our method achieves the best performance across all evaluation metrics.

\subsubsection{Semantic Segmentation}

To ensure a fair comparison, we adopt the same training configurations. For the semi-supervised landslide segmentation method Semi-LM \cite{ran2025semi}, we use the same labeled data and data splits as our method. In particular, we divide the dataset into training and test sets at an 8:2 ratio and consider labeled data ratios of 1/4 and 1/8 within the training set. All data split files are available in our open-source repository. For task-specific landslide segmentation methods, including 3D SCAM\cite{ji2020landslide} and PF-SAM\cite{11489313}, we use the same amount of labeled data to ensure a fair comparison. Since there has been little exploration of semi-supervised and label-efficient learning approaches in landslide studies, we additionally compare our method with several state-of-the-art general-purpose semi-supervised learning methods, including FixMatch \cite{sohn2020fixmatch}, UniMatch \cite{yang2023revisiting}, AllSpark \cite{Wang_2024_CVPR} and  UniMatch\_v2\cite{yang2025unimatch}. All these methods are trained with the same data configuration as our proposed method.

\begin{table*}[!t]
\caption{Comparison of semantic segmentation results with sota methods on bijie dataset. the best results are highlighted in bold.}
\label{tab: bijie}
\centering
\footnotesize
\begin{tabular}{ c |c c c c c c }
		\toprule
		\textbf{Ratio} & \textbf{Method} &\textbf{mPrecision (\%)} & \textbf{mRecall (\%)} & \textbf{mF1 (\%) } & \textbf{mIoU (\%) } & \textbf{Kappa (\%)}\\
		\midrule
		 \multirow{7}{*}{1/8} &  FixMatch \cite{sohn2020fixmatch} &  75.74 & 82.63 & 79.04 & 66.37 &  56.93 \\ 
           ~ &  UniMatch \cite{yang2023revisiting}  &  76.42 & 82.79 & 79.48 & 68.73 &  58.26 \\
          ~ & AllSpark \cite{Wang_2024_CVPR} &  78.81 & 84.60 &  81.60 & 71.89 &  62.80  \\
          ~ & UniMatch\_v2\cite{yang2025unimatch} &  84.53 & 89.18 &  86.79 & 77.90 &  73.26  \\
          ~ &  3D SCAM\cite{ji2020landslide}  &  70.66 & 79.77 &  74.94 & 64.74 &  48.56  \\
             ~ &  Semi-LM\cite{ran2025semi}  &  78.80 & 85.98 &  82.23 & 73.62 &  63.99  \\
          ~ &  PF-SAM\cite{11489313}  &  82.35 & 85.37 &  83.83 & 72.28 &  67.28  \\
           ~ &  \textbf{LandslideFM (Ours)}  &  \textbf{86.02} & \textbf{92.05} &  \textbf{88.93} & \textbf{80.47} &  \textbf{77.34}
          \\
         \midrule
		 \multirow{7}{*}{1/4} &  FixMatch \cite{sohn2020fixmatch} & 77.71 & 83.99 &  80.73 & 69.53 &  60.96 \\
           ~ &  UniMatch \cite{yang2023revisiting} &  77.83 & 83.43 &  80.53 & 69.51 &  60.57 \\
            ~ & AllSpark \cite{Wang_2024_CVPR} &  80.87 & 86.59 &  83.63 & 74.96 &  66.95  \\
           ~ & UniMatch\_v2\cite{yang2025unimatch} &  84.96 & 90.57 &  87.68 & 79.32 & 75.01  \\
          ~ &  3D SCAM\cite{ji2020landslide}  &  72.36 & 80.27 &  76.11 & 67.27 &  51.33  \\
          ~ &  Semi-LM\cite{ran2025semi}  &  80.69 & 87.31 &  83.87 & 75.24 &  67.27  \\
          ~ &  PF-SAM\cite{11489313}  &  84.10 & 86.69 &  85.38 & 77.17 &  70.45 \\
         ~ &  \textbf{LandslideFM (Ours)}  &  \textbf{86.51} & \textbf{92.94} &  \textbf{89.61} & \textbf{81.43} &  \textbf{78.62} 
          \\
		\bottomrule
	\end{tabular}
\end{table*}

\begin{table*}[!t]
\caption{Comparison of semantic segmentation results with sota methods on GVLM dataset. the best results are highlighted in bold.}
\label{tab: gvlm}
\centering
\footnotesize
\begin{tabular}{ c |c c c c c c }
		\toprule
		\textbf{Ratio} & \textbf{Method} &\textbf{mPrecision (\%)} & \textbf{mRecall (\%)} & \textbf{mF1 (\%) } & \textbf{mIoU (\%) } & \textbf{Kappa (\%)}\\
		\midrule
		 \multirow{7}{*}{1/8} &  FixMatch \cite{sohn2020fixmatch} &  75.68 & 80.42 & 77.98 & 66.23 &  55.51 \\ 
           ~ &  UniMatch \cite{yang2023revisiting}  &  74.43 & 81.20 & 77.67 & 66.57 &  54.58 \\ 
           ~ & AllSpark \cite{Wang_2024_CVPR} &  76.20 & 81.62 & 78.82 & 68.45 &  57.21 \\ 
          ~ & UniMatch\_v2\cite{yang2025unimatch} &  90.58 & 87.94 & 89.24 & 80.54 &  78.14 \\ 
          ~ &  3D SCAM\cite{ji2020landslide}  &  72.38 & 77.72 & 74.96 & 66.09 &  49.50 \\ 
             ~ &  Semi-LM\cite{ran2025semi}  &  85.22 & 85.78 & 85.50 & 77.32 &  70.95 \\ 
          ~ &  PF-SAM\cite{11489313}  &  87.10 & 86.84 & 86.97 & 76.08 &  73.30 \\ 
           ~ &  \textbf{LandslideFM (Ours)}  & \textbf{93.88}  & \textbf{91.89} & \textbf{92.87} & \textbf{87.33} &  \textbf{85.72} \\ 
         \midrule
		 \multirow{7}{*}{1/4} &  FixMatch \cite{sohn2020fixmatch} &  75.95 & 81.74 & 78.74 & 68.78 &  57.02 \\ 
           ~ &  UniMatch \cite{yang2023revisiting} &  75.85 & 81.51 & 78.58 & 69.85 &  56.76 \\
              ~ & AllSpark \cite{Wang_2024_CVPR} &  77.29 & 82.66 & 79.88 & 70.15 &  59.40 \\
           ~ & UniMatch\_v2\cite{yang2025unimatch} &  91.25 & 89.90 & 90.57 &  82.67 &  80.74 \\ 
          ~ &  3D SCAM\cite{ji2020landslide}  &  74.29 & 80.13 & 77.10 & 66.99 &  53.69 \\ 
             ~ &  Semi-LM\cite{ran2025semi}  &  86.82 & 88.40 &   87.60 &  77.72 & 74.91 \\ 
          ~ &  PF-SAM\cite{11489313}  &  89.71 & 89.25 & 89.48 & 81.23 &  78.73 \\ 
            ~ &  \textbf{LandslideFM (Ours)}  &  \textbf{94.09} & \textbf{93.64} & \textbf{93.86} & \textbf{88.93} &  \textbf{87.73} \\ 
		\bottomrule
	\end{tabular}
\end{table*}

The experimental results are reported in Tables \ref{tab: bijie} and \ref{tab: gvlm}. Our method achieves the best performance across all evaluation metrics on both the BJL and GVLM datasets, including mPrecision, mRecall, mF1, mIoU, and Kappa. These consistent improvements indicate more comprehensive and balanced segmentation performance, with fewer false positives and false negatives and greater robustness. It is worth noting that, since semantic segmentation performs pixel-level classification of both landslide and background classes, all reported metrics are averaged over the two classes. In contrast, the classification and detection tasks evaluate only the landslide class.

\begin{figure*}[!t]
\centering
\includegraphics[width=7in]{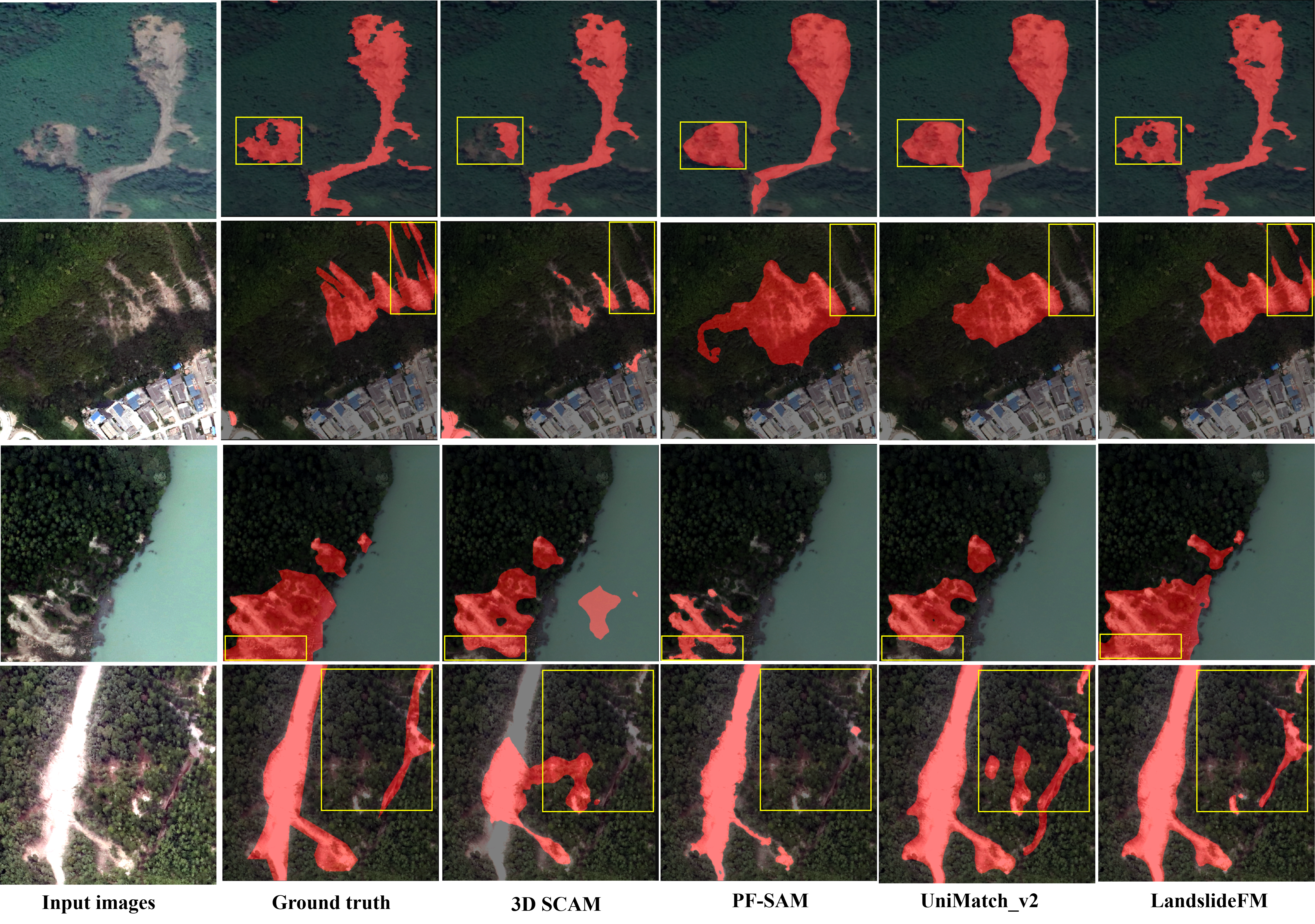}
\caption{Qualitative results with different SOTA landslide methods on the Landslide datasets.}
\label{fig: seg}
\end{figure*}

We further compare the qualitative segmentation results of different landslide segmentation methods, as shown in Fig. \ref{fig: seg}. The selected images cover diverse scenarios, including landslides in forested areas and those near urban areas. Yellow rectangles highlight regions where other SOTA methods produce segmentation errors, showing the accurate segmentation results achieved by our method in the same regions for comparison. Previous methods generally identify the approximate locations of landslides but struggle with pixel-level segmentation under complex environmental conditions. Specifically, as shown in the first row of Fig. \ref{fig: seg}, most methods detect the main landslide regions but produce noticeable errors in the smaller areas highlighted by the yellow rectangles. In the second row, other methods miss a substantial portion of the landslide above the residential area, potentially leading to serious consequences in practical applications. Similar segmentation errors and missed detections are observed in the third and fourth rows. In contrast, LandslideFM achieves superior segmentation performance, particularly in delineating fine-grained structures and accurately identifying landslide boundaries.

\vspace{-0cm}

\subsection{Intelligent Analysis on the LandslideVL Benchmark}

This section evaluates the intelligent analysis capabilities of LandslideFM on our newly constructed LandslideVL benchmark. Given the absence of domain-specific baselines for landslide scene interpretation, we evaluate our model primarily against representative general-purpose multimodal large language models, such as Qwen2.5-VL-3B\cite{bai2025qwen2.5vl}, Qwen2.5-VL-7B\cite{bai2025qwen2.5vl}, Qwen3-VL-8B\cite{bai2025qwen3} and Kimi-VL-A3B\cite{team2026kimi}. We first evaluate each model across multiple performance metrics using the structured text annotations in the LandslideVL dataset.
The experimental results are presented in Table \ref{tab: LandslideVL_structured}, while the average results across all metrics are reported in Table \ref{tab: LandslideVL_structured2}. In the table, P and R represent precision and recall, respectively. The experimental results in Table \ref{tab: LandslideVL_structured} show that LandslideFM achieves substantial improvements over other SOTA VLMs. Specifically, its precision for the landslide\_exists field increases by 24.9\%. For other fields, such as has\_building, its precision is only slightly lower than that of the other VLMs, but its recall improves substantially, being 40.12\% higher than that of the second-best model. This is because the other VLMs have not been trained with domain-specific knowledge of RS landslides, resulting in severe missed detections for certain fields. Our model also achieves significant improvements in terms of F1 score. Furthermore, the experimental results in Table \ref{tab: LandslideVL_structured2} show that our model achieves the best average performance in all fields. In particular, its Recall is 27.31\% higher than that of the second-best model. It is worth noting that the F1 score in Table \ref{tab: LandslideVL_structured2} is obtained by averaging the F1 scores in Table \ref{tab: LandslideVL_structured}, rather than calculated from the precision and recall values in Table \ref{tab: LandslideVL_structured2}.

\begin{table*}[!t]
    \caption{Comparison of structured text generation results across different binary scene attributes on the LandslideVL dataset with SOTA methods. The best results are highlighted in bold.}
  \label{tab: LandslideVL_structured} 
    \centering

    \resizebox{\textwidth}{!}{%
    \begin{tabular}{l*{15}{c}}
        \toprule
        \textbf{Models}
        & \multicolumn{3}{c}{\textbf{landslide\_exists}}
        & \multicolumn{3}{c}{\textbf{has\_building}}
        & \multicolumn{3}{c}{\textbf{has\_road}}
        & \multicolumn{3}{c}{\textbf{has\_water}}
        & \multicolumn{3}{c}{\textbf{has\_vegetation}} \\

        \cmidrule(lr){2-4}
        \cmidrule(lr){5-7}
        \cmidrule(lr){8-10}
        \cmidrule(lr){11-13}
        \cmidrule(lr){14-16}

        & P & R & F1
        & P & R & F1
        & P & R & F1
        & P & R & F1
        & P & R & F1 \\

        \midrule

        Qwen2.5-VL-3B\cite{bai2025qwen2.5vl}
        & 59.35 & 98.54 & 74.08
        & \textbf{99.79} & 55.54 & 71.36
        & \textbf{98.13} & 37.01 & 53.75
        & \textbf{99.15} & 16.84 & 28.79
        & 99.21 & 75.14 & 85.51 \\

        Qwen2.5-VL-7B\cite{bai2025qwen2.5vl}
        & 70.74 & 98.56 & 82.36
        & 97.96 & 27.43 & 42.86
        & 96.28 & 41.00 & 57.51
        & 93.92 & 44.74 & 60.61
        & 99.46 & 84.69 & 91.48 \\

        Qwen3-VL-8B\cite{bai2025qwen3}
        & 72.81 & 97.44 & 83.34
        & 98.76 & 46.19 & 62.94
        & 97.80 & 53.37 & 69.06
        & 95.86 & 50.66 & 66.29
        & 99.04 & 87.93 & 93.15 \\

         Kimi-VL-A3B \cite{team2026kimi}
        & 56.37 & 98.84 & 71.79
        & 99.64 & 38.97 & 56.03
        & 97.04 & 67.37 & 79.53
        & 86.36 & 23.24 & 36.63
        & 99.43 & 99.74 & 99.58 \\
        \midrule
         
        \multirow{2}{*}{\textbf{LandslideFM (Ours)}}
  
        & \textbf{97.71} & \textbf{98.88} & \textbf{98.29}
        & 92.08 & \textbf{95.66} & \textbf{93.83}
        & 89.04 & \textbf{91.96} & \textbf{90.48}
        & 87.87 & \textbf{85.79} & \textbf{86.82}
        & \textbf{99.51} & \textbf{99.87} & \textbf{99.69} \\

        & $\uparrow$ 24.90 & $\uparrow$ 0.04 & $\uparrow$ 14.95
        & $\downarrow$ 7.71 & $\uparrow$ 40.12 & $\uparrow$ 22.47
        & $\downarrow$ 9.09 & $\uparrow$ 24.59  & $\uparrow$ 10.95
        & $\downarrow$ 11.28 & $\uparrow$  35.13 & $\uparrow$ 20.53
        & $\uparrow$ 0.05 & $\uparrow$ 0.13 & $\uparrow$ 0.11\\
        
        \bottomrule
    \end{tabular}%
    }
\end{table*}

\begin{table}[h]
\caption{Comparison of the average performance metrics across all fields for structured text generation with SOTA methods on the LandslideVL dataset.}
\label{tab: LandslideVL_structured2}
\centering
\begin{tabular}{c|cccc}
    \toprule
    \textbf{~} & \textbf{Models} & \textbf{mPrecision} & \textbf{mRecall} & \textbf{mF1} \\
    \midrule
    \multirow{4}{*}{Average}
        & Qwen2.5-VL-3B\cite{bai2025qwen2.5vl} & 91.13 & 56.61 & 62.70 \\
        & Qwen2.5-VL-7B\cite{bai2025qwen2.5vl} & 91.67 & 59.28 & 66.96 \\
        &  Qwen3-VL-8B\cite{bai2025qwen3} & 92.85 & 67.12 & 74.96 \\
        & Kimi-VL-A3B \cite{team2026kimi} & 87.77 & 65.63 & 68.71 \\
        &  \textbf{LandslideFM (Ours)}& \textbf{93.24} & \textbf{94.43} & \textbf{93.82} \\ 
    \bottomrule
\end{tabular}
\end{table}

In addition to structured text validation, we also focused on evaluating the performance of each model in the detailed analysis of landslide images described in this paper. The experimental results are presented in Table \ref{tab: LandslideVL_text}. LandslideFM substantially outperforms the other VLMs in generating detailed textual descriptions for intelligent landslide analysis, achieving a BLEU score of 13.16\%, compared with 0.57\% for the best-performing baseline. The precision (P), recall (R), and F1 scores in Table \ref{tab: LandslideVL_text} are computed separately for each image and then arithmetically averaged across all images, corresponding to macro-averaging (i.e., image-wise averaging). The reported F1 score is marginally lower than the average of the reported P and R values. Further details of the calculation procedure are available in our code repository.

\begin{table*}[!t]
\caption{Comparison of detailed text generation results with SOTA methods on LandslideVL. The best results are highlighted in bold.}
\label{tab: LandslideVL_text}
\centering
\footnotesize
\begin{tabular}{ c | c c c c c c c}
		\toprule
		\textbf{Method} & \textbf{BLEU} & $\textbf{ROUGE-L}_P$ & $\textbf{ROUGE-L}_R$ & $\textbf{ROUGE-L}_{F1}$  & $\textbf{P}_{WO}$ & $\textbf{R}_{WO}$ & $\textbf{F1}_{WO}$ \\
		\midrule
		Qwen2.5-VL-3B\cite{bai2025qwen2.5vl}  & 0.12 & 4.64  & 27.04 &  7.87 & 11.06 & 37.92 &  17.02  \\
        Qwen2.5-VL-7B\cite{bai2025qwen2.5vl} &   0.24 &  4.29 & 30.20 & 7.44  & 9.64 & 41.20 & 15.55 \\
        Qwen3-VL-8B\cite{bai2025qwen3}  & 0.57 &  5.17 & 30.92 & 8.63 & 12.79 & 41.83 & 19.27 \\
        Kimi-VL-A3B \cite{team2026kimi} & 0.17 & 4.12 & 30.82 & 7.27 & 9.39 & 41.07 & 15.08\\ 
        \textbf{LandslideFM (Ours)} &  \textbf{13.16}  &  \textbf{35.44} & \textbf{32.83} &  \textbf{32.50} & \textbf{45.40} & \textbf{42.05} & \textbf{41.94} \\
		\bottomrule
	\end{tabular}
\end{table*}

\begin{figure*}[!t]
\centering
\includegraphics[width=7.2 in]{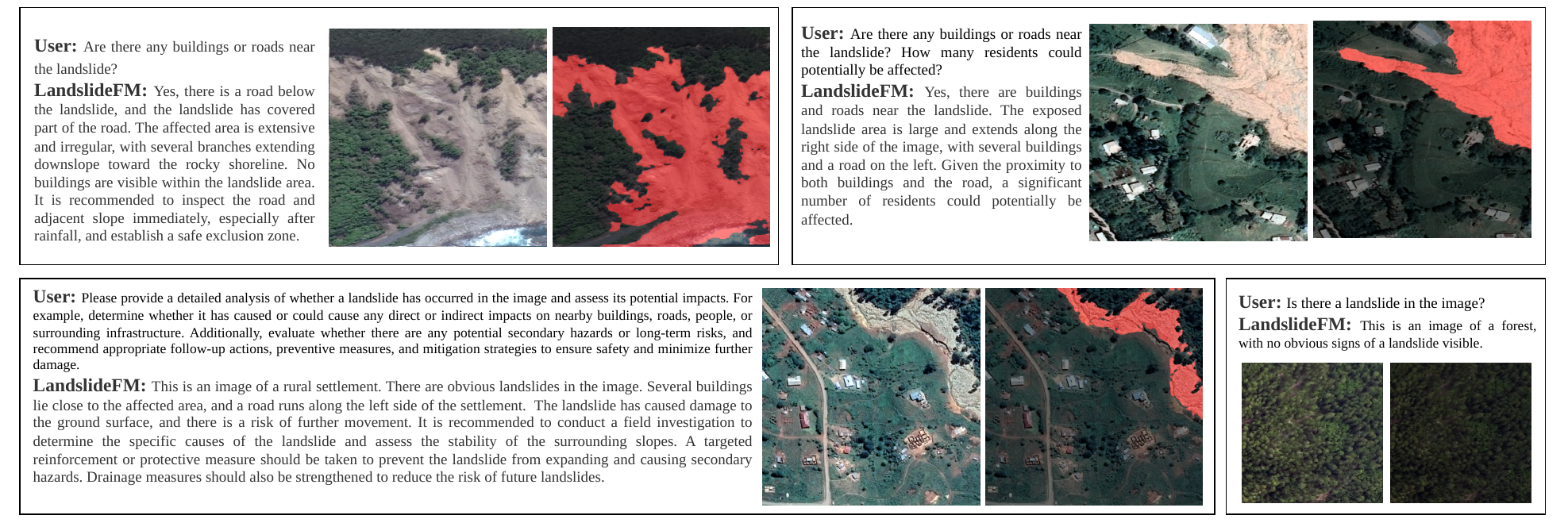}
\caption{Several real-world examples of intelligent landslide image analysis.}
\label{fig: analysis}
\end{figure*}

Figure \ref{fig: analysis} presents real-world examples of LandslideFM’s intelligent analysis of landslide imagery. These examples illustrate its ability to analyze landslide sites across diverse scenarios and provide practical recommendations for disaster prevention and mitigation.

\vspace{-0cm}

\subsection{Ablation Study}
This subsection presents ablation studies of key components and parameters of LandslideFM, focusing on SAGT-based semi-supervised pre-training and intelligent landslide analysis.

\subsubsection{Ablation of the SAGT}

This subsection presents ablation experiments on SAGT semi-supervised pre-training. Specifically, we compare the original pre-training method without any SAGT components (Without SAGT) with a variant of SAGT that does not progressively exploit disagreement regions (No Graduated). In the No Graduated setting, disagreement regions between the VFM and the teacher model are directly discarded. The experiments are conducted using a labeled data ratio of $1/8$. As shown in Tab.\ref{tab: Ablation of SAGT}, SAGT achieves substantial improvements across all evaluation metrics compared with both No Graduated and Without SAGT, demonstrating the effectiveness of SAGT. In the next subsection, we further conduct an ablation study on the graduated trust weights $w_i$.

\begin{table}[!t]
\caption{Ablation of the SAGT semi-supervised pre-training. \label{tab: Ablation of SAGT}}
\centering
\footnotesize
\begin{tabular}{ c |c | c |c |c }
		\toprule
		    \textbf{Dataset} & \textbf{Network} &\textbf{mF1} &\textbf{mIoU} &\textbf{kappa}\\
		\midrule
        \multirow{4}{*}{BJL} & Without SAGT  & 78.46 & 69.35 & 55.76 \\
        \cmidrule{2-5}
        ~ &  No Graduated & 84.84 & 76.40 & 68.60   \\
        \cmidrule{2-5}
        ~ &  Full SAGT & \textbf{88.93} & \textbf{80.47} & \textbf{77.34}   \\
        \cmidrule{1-5}
          \multirow{4}{*}{GVLM} & Without SAGT  & 85.41 & 77.29 & 70.72 \\
        \cmidrule{2-5}
        ~ &  No Graduated & 90.35 & 83.45 & 79.20   \\
        \cmidrule{2-5}
         ~ &  Full SAGT & \textbf{92.87} & \textbf{87.33} & \textbf{85.72}  \\
		\bottomrule
	\end{tabular}
\end{table}

\subsubsection{Ablation of the Graduated Trust Weights $w_i$}
\begin{figure}[!t]
\centering
\includegraphics[width=3.5in]{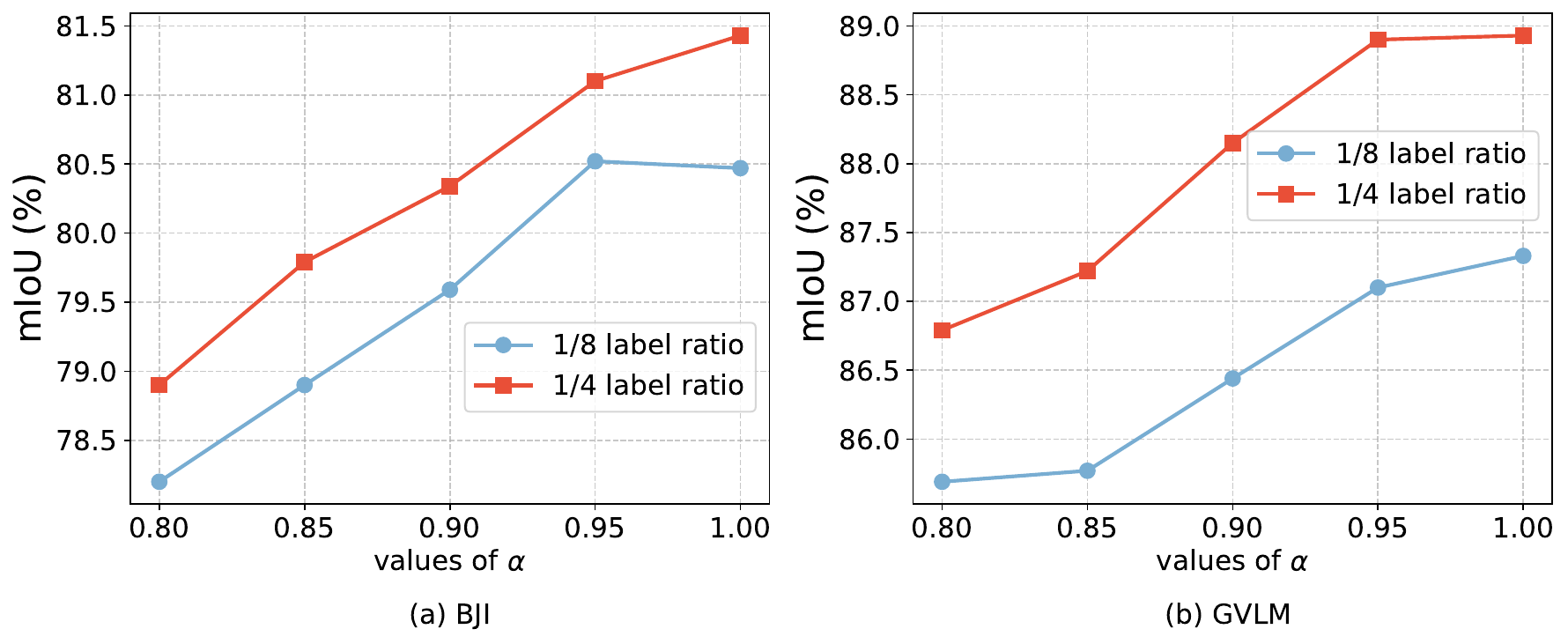}
\vspace{-0.5cm}
\caption{Ablation of the $\alpha$ of the graduated trust weights $w_i$.}
\label{fig: alpha}
\vspace{-0.5 cm}
\end{figure}

In Eq. \ref{eq: w_i}, $w_i = \alpha$ corresponds to cases where the teacher model and the VFM produce reliable and mutually consistent predictions. We therefore assign a relatively high value to $\alpha$. Figure \ref{fig: alpha} shows how semantic segmentation mIoU varies with $\alpha$. The experimental results show that mIoU increases as the value of $\alpha$ increases, reaching its maximum when $\alpha$ is between 0.95 and 1. This observation is consistent with our analysis: a higher weight $w_i$ is assigned when the predictions of the VFM and the teacher model agree.

\begin{figure}[!t]
\centering
\includegraphics[width=3.5in]{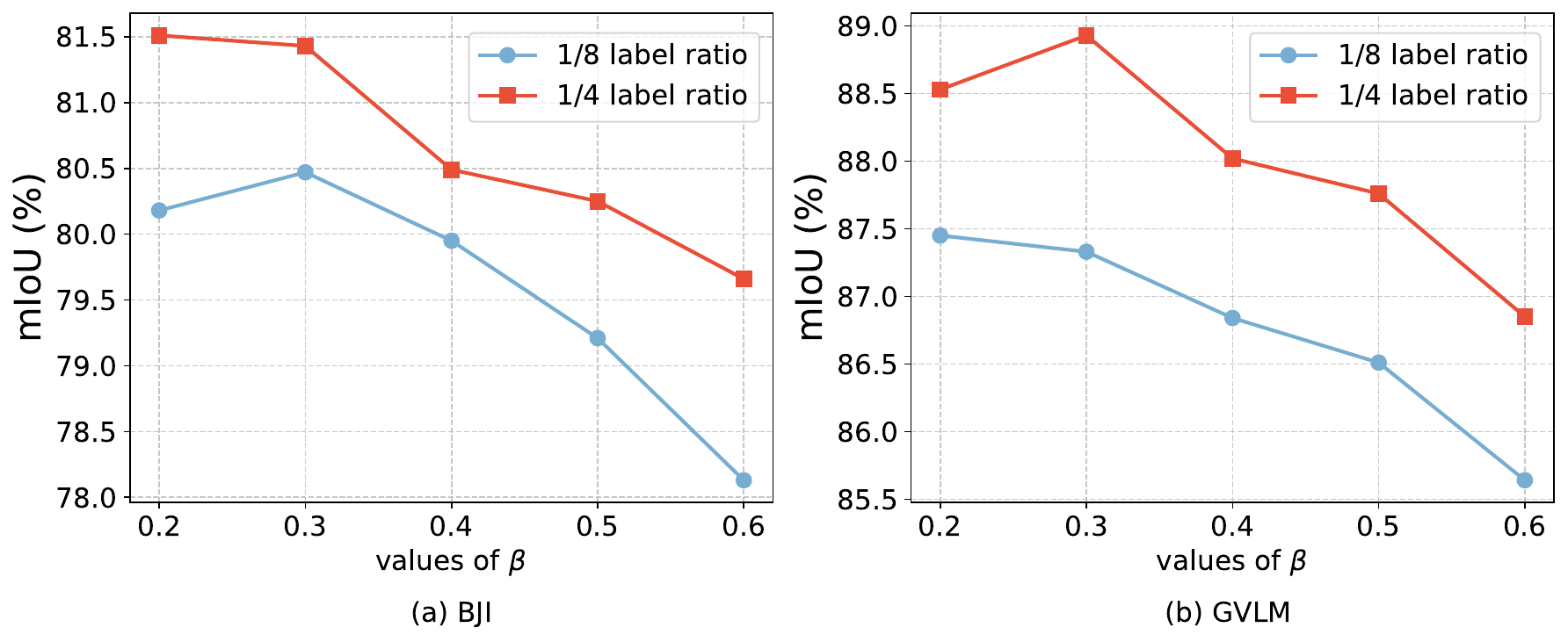}
\vspace{-0.5cm}
\caption{Ablation of the $\beta$ of the graduated trust weights $w_i$.}
\label{fig: beta}
\vspace{-0.5 cm}
\end{figure}

In case of  $w_i = \beta$, both the VFM and the teacher model produce high-confidence pseudo-labels, yet they provide conflicting predictions. This case is only slightly better than the case in which neither model produces a reliable pseudo-label. We therefore assign a relatively low value to $\beta$. As shown in Fig. \ref{fig: beta}, semantic segmentation mIoU decreases as $\beta$ increases. Based on these results, we set $\beta= 0.3$ in the weighting scheme.

\begin{figure}[!t]
\centering
\includegraphics[width=3.5in]{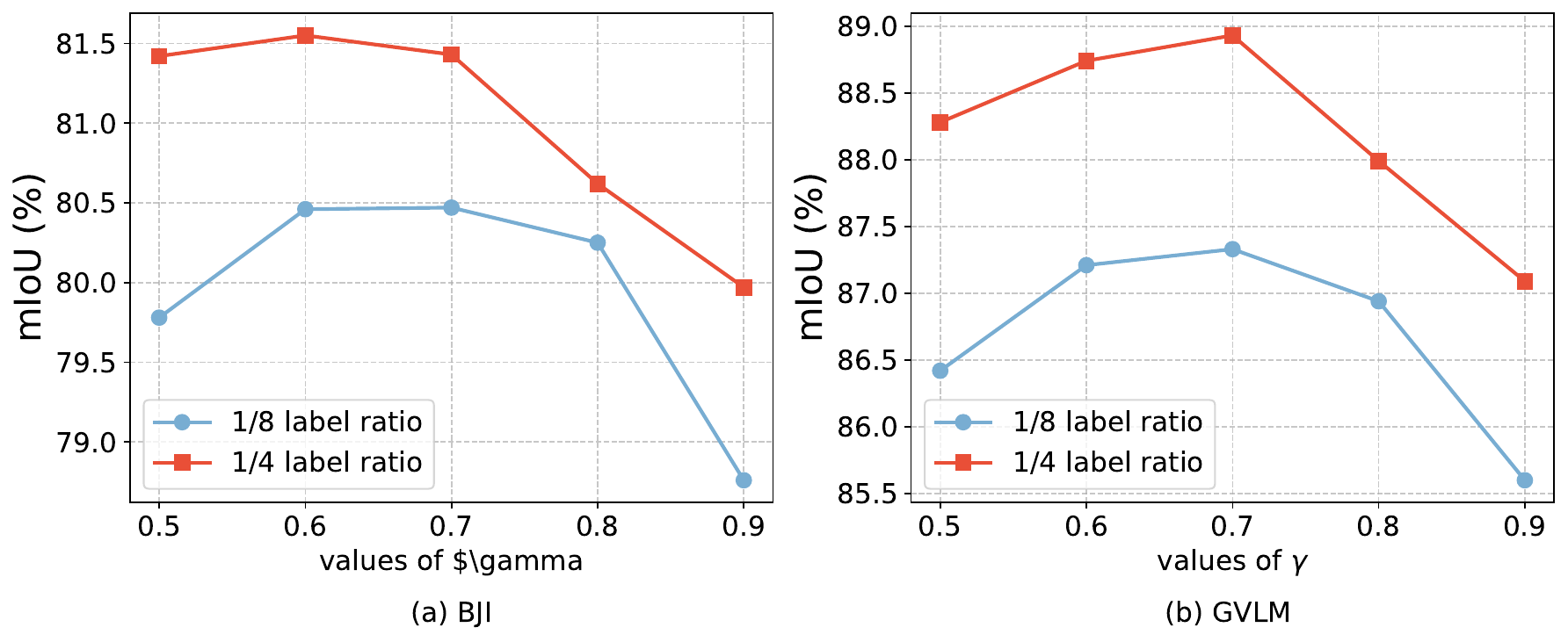}
\vspace{-0.7cm}
\caption{Ablation of the $\gamma$ of the graduated trust weights $w_i$.}
\label{fig: gamma}
\end{figure}

\begin{figure}[!t]
\centering
\includegraphics[width=3.5in]{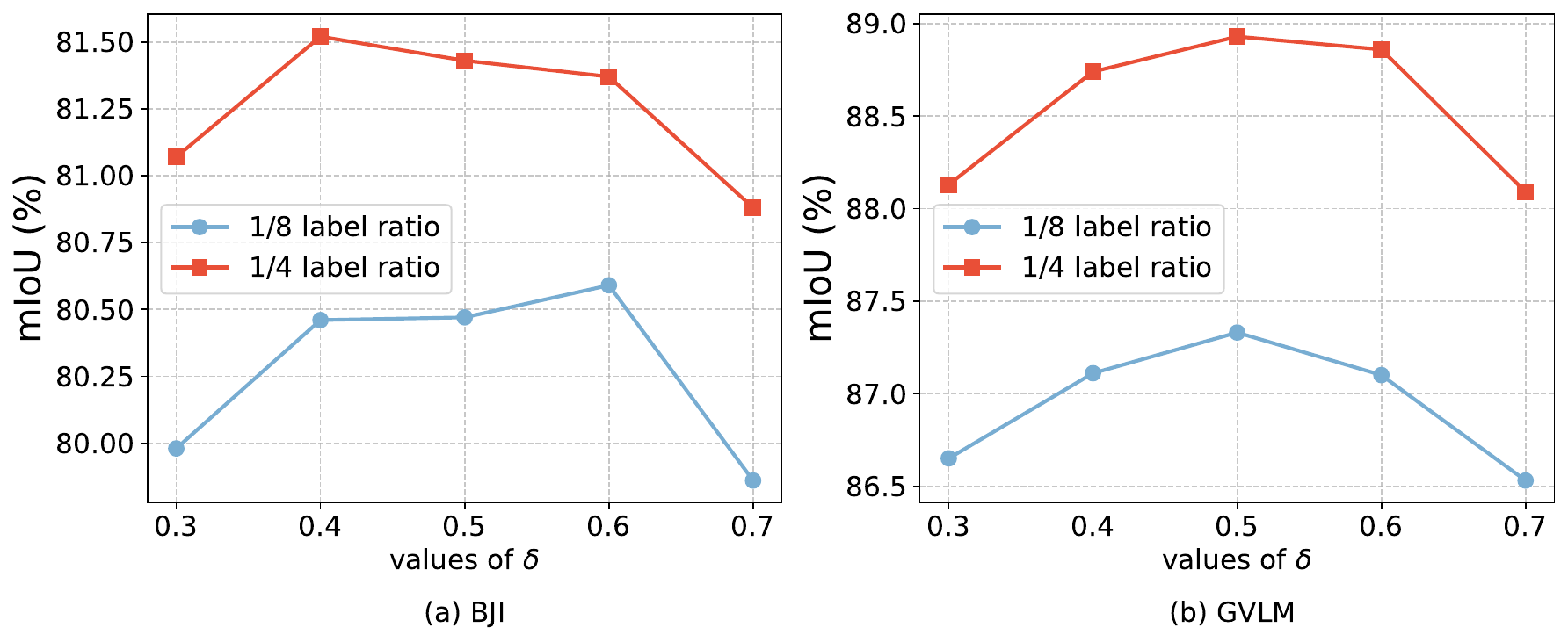}
\vspace{-0.7cm}
\caption{Ablation of the $\delta$ of the graduated trust weights $w_i$.}
\label{fig: delta}
\end{figure}

The cases of  $w_i = \gamma$ or $w_i = \delta$ imply that only one of the pseudo-labels generated by the VFM or the teacher model is reliable. In this scenario, the values of $\gamma$ and $\delta$ are set lower than $\alpha$. Furthermore, since the VFM is generally more reliable than the teacher model, $\gamma$ is set slightly higher than $\delta$. We investigate how the semantic segmentation mIoU varies with different values of $ \gamma$ and $\delta$, as shown in Fig. \ref{fig: gamma} and Fig. \ref{fig: delta}. The experimental results indicate that the highest mIoU is achieved when $\gamma$ ranges from 0.6 to 0.7 and $\delta$ ranges from 0.4 to 0.6. Based on these findings, we finally set $\gamma = 0.7$ and $\delta = 0.5$.

\subsubsection{Intelligent Landslide Analysis Training Process}
\begin{figure}[h]
\centering
\includegraphics[width=3.5in]{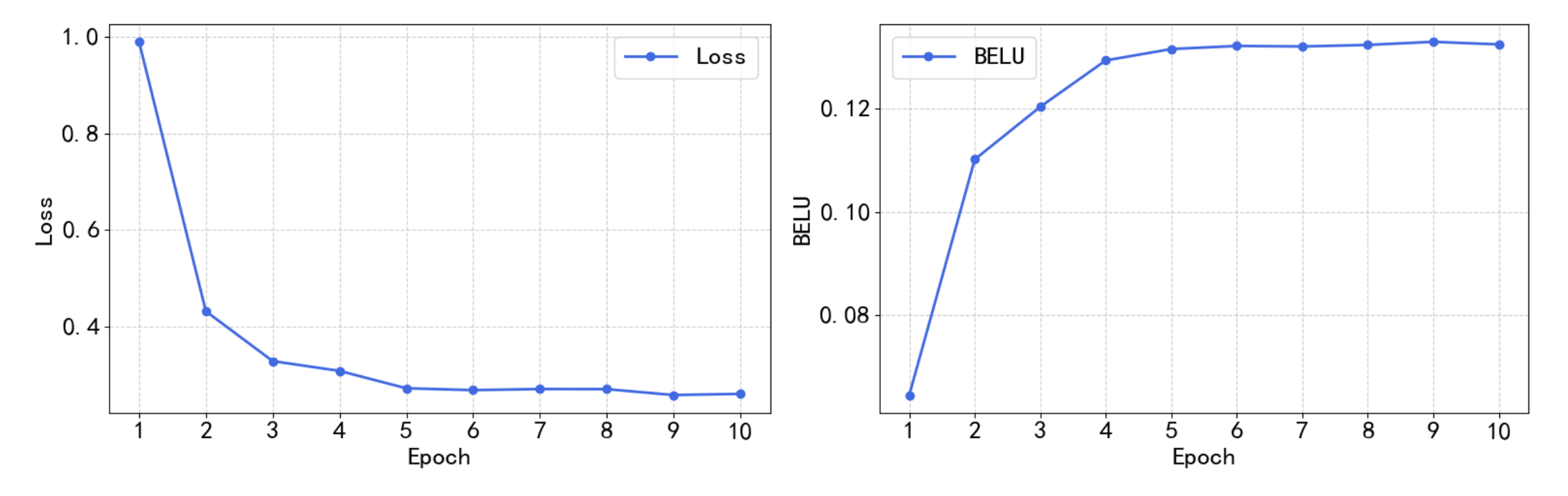}
\vspace{-0.5cm}
\caption{Evolution of LandslideFM’s landslide analysis capabilities during fine-tuning on LandslideVL.}
\label{fig: iterative}
\vspace{-0.5 cm}
\end{figure}

In this subsection, we visually present how intelligent landslide analysis capabilities evolve across iterations on the LandslideVL benchmark. As shown in Fig. \ref{fig: iterative}, the loss decreases rapidly while the BLEU score increases sharply as training progresses. The training process converges to a stable state within approximately five epochs, demonstrating that LandslideFM successfully acquires intelligent landslide analysis capabilities through fine-tuning on the LandslideVL dataset.

\section{Conclusion}
\label{sec: Conclusion}
Our study addresses the challenge of unifying visual understanding and intelligent analysis for landslide detection through the proposed LandslideFM. LandslideFM employs SAGT-based semi-supervised pretraining to develop a powerful, unified visual backbone for multiple downstream vision tasks and to provide visual support for intelligent landslide analysis. Furthermore, we introduce LandslideVL, the first benchmark for intelligent landslide analysis, designed to support the training and fine-tuning of VLMs. Compared with other SOTA methods, our model achieves superior performance across multiple visual tasks and, unlike most previous methods, demonstrates strong intelligent analysis capabilities. Specifically, our model significantly outperforms several state-of-the-art general-purpose vision-language models, including Qwen3-VL and Kimi-VL-A3B, on the LandslideVL benchmark. Nevertheless, as the first foundation model to unify visual understanding and intelligent analysis for landslides, LandslideFM still faces several challenges. For instance, the potential of integrating large AI models with complex spectral imaging for landslide analysis within a unified framework remains largely unexplored. Future work will investigate how large AI models can be effectively used to address the challenges associated with complex spectral imaging. Finally, as the first study to introduce both a foundation model and a dedicated benchmark for landslide analysis, we hope that LandslideFM and LandslideVL will serve as valuable resources and help advance future research in this emerging field.

\bibliographystyle{IEEEtran}
\bibliography{ref}

\end{document}